# Extracting Epistatic Interactions in Type 2 Diabetes Genome-Wide Data Using Stacked Autoencoder

Basma Abdulaimma, Paul Fergus, Carl Chalmers

Liverpool John Moores University, Byrom Street, Liverpool, L3 3AF, UK

**Abstract**— Type 2 Diabetes is a leading worldwide public health concern, and its increasing prevalence has significant health and economic importance in all nations. The condition is a multifactorial disorder with a complex aetiology. The genetic determinants remain largely elusive, with only a handful of identified candidate genes. Genome wide association studies (GWAS) promised to significantly enhance our understanding of genetic based determinants of common complex diseases. To date, 83 single nucleotide polymorphisms (SNPs) for type 2 diabetes have been identified using GWAS. Standard statistical tests for single and multi-locus analysis such as logistic regression, have demonstrated little effect in understanding the genetic architecture of complex human diseases. Logistic regression is modelled to capture linear interactions but neglects the non-linear epistatic interactions present within genetic data. There is an urgent need to detect epistatic interactions in complex diseases as this may explain the remaining missing heritability in such diseases. In this paper, we present a novel framework based on deep learning algorithms that deal with non-linear epistatic interactions that exist in genome wide association data. Logistic association analysis under an additive genetic model, adjusted for genomic control inflation factor, is conducted to remove statistically improbable SNPs to minimize computational overheads. SNPs generated with a p-value threshold of $10^{-2}$ are considered, resulting in epistatic analysis in the remaining 6609 SNPs. Latent representations are extracted using a stacked autoencoder model and used to initialize a deep learning classifier for binary classification tasks. The performance of the proposed stacked autoencoder model is evaluated and benchmarked against traditional machine learning models. The findings show that using 2500 hidden units from our stacked autoencoder it is possible to obtain (AUC=92.89%, Sens=90.87%, Spec=80.53%, Logloss= 45.82%, Gini=85.78%, and MSE=11.62%). Deterioration in classification accuracy (AUC=89.38%) is observed using 700 hidden units.

─ ─ ─ ─ ─ ─ ─ ─ ─  ◆  ─ ─ ─ ─ ─ ─ ─ ─ ─

## 1 Introduction

TYPE 2 Diabetes (T2D) is a multifactorial disorder, caused by the convergence of genetics, environment, and a sedentary lifestyle [1]. There is strong evidence that genetic factors play a significant role in T2D susceptibility [2]. Twin studies have shown that the concordance rate of T2D in monozygotic twins is approximately 70% compared with 20% to 30% in dizygotic twins [3]. As such, an in-depth investigation into T2D genetic data is needed. This may help researchers and professionals gain a better understanding of T2D aetiology.

With the availability of less expensive genotyping technologies [4], genome wide association studies (GWAS) have seen widespread use within genetic research. In recent years, GWAS have succeeded in identifying genetic variants that show evidence of increased susceptibility to a wide range of complex diseases, including Type 2 Diabetes, Schizophrenia, Epilepsy, Obesity, Cardiovascular Disease, and Hypertension [5], [6].

GWAS have also been used to detect the main genetic effects associated with the phenotype (disease trait) in case-control studies using single-locus statistical tests. These type of studies explore each single nucleotide polymorphisms (SNPs) separately [7]. Most genetic variants identified so far explain a relatively small proportion of the overall heritability, leaving an open question about how remaining, missing heritability can be better explained [8], [9]. The underlying cause of complex human diseases does not rely on single genetic variations but rather contribution of multiple interacting genetic loci [10], [11], [12]. The issue with GWAS, however is that this approach fails to find the non-linear relationships between genotypes and the phenotype under investigation. Standard multi-variable statistical approaches, such as logistic regression, are more suited to capturing linear interactions between genotypes and the phenotype for simpler diseases like cystic fibrosis which is known to only have one associated SNP.

Existing studies using GWAS data have focused on the use of machine learning algorithms [13], [14], [15], [16]. Machine Learning techniques can model complex relationships and interactions between features (SNPs) and their association to the phenotype. Ensemble methods like the random forest (RF) algorithm have been broadly applied for genomic data analysis to detect SNP correla-

─────────────────

- *Basma Abdulaimma is with the Department of Computer Science, Liverpool John Moores University, Byrom Street, Liverpool, L3 3AF, UK. E-mail: B.T.Abdulaimma@2015.ljmu.ac.uk.*
- *Paul Fergus is with the Department of Computer Science, Liverpool John Moores University, Byrom Street, Liverpool, L3 3AF, UK. E-mail: P.Fergus@ljmu.ac.uk.*
- *Carl Chalmers is with the Department of Computer Science, Liverpool John Moores University, Byrom Street, Liverpool, L3 3AF, UK. E-mail: c.chalmers@ljmu.ac.uk.*

tions [13], disease risk prediction [14], and feature selection [15]. Support Vector Machines (SVMs) have been used for the detection of gene-gene interactions [16], and disease classification [17]. While, Artificial Neural Networks (ANNs) have been utilized to detect SNP correlations as demonstrated in [18], [19]. Although, these types of machine learning algorihm are competent in handling complex correlations and interactions among a small number of features, they do not scale to larger numbers of SNPs which is the case when using GWAS data (genotypes of almost one million SNPs and thousands of samples). The main issues traditional machine learning algorithms suffer with include multicollinearity [20], and the curse of dimensionality [21]. Therefore, an alternative approach to model high-dimensional GWAS data and epistatic interactions between SNPs is needed. Using unsupervised deep learning (DL) algorithm seems appealing since it exhibits the potential to deal with big data and the detection of complex features and associated interactions.

In this paper, we consider the application of deep learning stacked autoencoders to model epistatic interactions between SNPs and fine-tune a fully connected multi-layer perceptron (MLP). The efficacy of the approach is evaluated by classifying observations as either case or control in a T2D GWAS data set.

To the best of our knowledge, this is the first study of its kind to combines unsupervised learning built on stacked autoencoders with an MLP classifier for the classification of T2Ds using GWAS data.

The remainder of this paper organized as follows. Section 2 provides details about the materials and methods used in this study. The results are presented in Section 3, while the findings are discussed in Section 4 before the paper is concluded in Section 5.

## 2 MATERIALS AND METHODS

### 2.1 Data Description

The Nurses' Health Study (NHS) and the Health Professionals Follow-up Study (HPFS) in T2D are used in this study (provided by the Genotypes and Phenotypes (dbGap) database) [25]. The NHS and HPFS cohorts are part of the Gene Environment Association Studies initiative (http://www.genevastudy.org) funded by the trans-NIH Genes, Environment, and Health Initiative (GEI). Participants were selected from those who provided a blood sample. Case participants were defined as those who reported themselves to be affected by T2D and confirmed by a medical record validation questionnaire. Control candidates were identified as those without diabetes. The Deoxyribonucleic acid (DNA) of selected candidates were genotyped at the Broad Centre for Genotyping and Analysis (CGA) using the Affymetrix Genome-Wide Human 6.0 array (Affymetrix is a DNA microarray technology that enables multiplex and parallel analysis of biological systems at the cell, protein, and gene level).

The final version of the dataset includes 6041 NHS and HPFS case-control subjects with genotype information across 909622 SNPs. The NHS participants consist of 1581 T2D cases and 1854 controls, and the HPFS subjects comprise 1232 T2D cases and 1374 controls. Participants in the NHS dataset are identified as Hispanic or non-Hispanic, and each belongs to one of four racial categories (White, African-American, Asia or Other). Participants are mainly White and non-Hispanic representing 97.4% of the NHS subjects. The HPFS participants belong to one of the four racial categories (White, African-American, Asia or Other). They are predominantly White representing 96% of the HPFS subjects.

### 2.2 Data Preprocessing

PLINK v1.07 and v1.9 [26] for Windows is used to conduct data quality control (QC) and preliminary analysis. PLINK is also utilized to merge the NHS and HPFS datasets (NHS and HPFS participants were genotyped using the Affymetrix Genome-Wide Human 6.0 array). Before QC, the 0 Chromosome was removed, the HapMap controls (44 NHS, 29 HPFS), non-T2D participants, i.e., other types of diabetes (65 NHS, 68 HPFS), and those belonging to ethnicity other than white (61 NHS, 103 HPFS) were excluded from the study. This study is restricted to white ancestry to reduce potential bias due to population stratification. The dataset was subjected to pre-established quality control protocols as recommended in [27]. In addition, the threshold measurements for quality control parameters were tuned to meet the requirements of the analysis presented in this study. Quality control assessments for individuals and genetic data are conducted separately.

Samples with discordant sex information (homozygosity rate of 0.2 for female, 0.8 for male) were identified resulting in 14 individuals being removed from the dataset. Samples with elevated missing data rates (genotype rate ≥ 0.05) and outlying heterozygosity rate (heterozygosity rate ±3 standard deviations from the mean) were identified resulting in 131 individuals being discarded from the analysis. Individuals with divergent ancestry were identified using the 2nd principal component score < 0.061 resulting in 51 individuals being removed. Identity-by-descent (IBD) was measured to remove duplicated or related individuals (IBD > 0.185). This resulted in eight individuals being excluded from the dataset. 101 individuals were removed due to missing genotype data rate of 0.05.

Genetic Markers (SNPs) that met any of the following criteria were discarded from the analysis. SNPs with excessive missing data rates were identified resulting in 29 SNPs being excluded. 116863 variants with missing genotype rate of 0.01 and 178004 variants with minor allele frequency (MAF) < 0.05 were removed. 2248 variants removed due to Hardy-Weinberg Equilibrium (HWE) with p-value < 0.001 in control samples. Following the QC steps, there were 5393 individuals (2481 cases, 2912 controls) and 608342 markers with a 0.961665 genotype rate in the remaining samples.

### 2.3 Association Analysis

In the current study, standard case-control association analysis is conducted in an unrelated, white racial sub-population to compare the frequency of alleles or genotypes at genetic marker loci (SNPs) between cases and controls of the merged version of Geneva NHS and HPFS Datasets. Association analysis is performed using PLINK v1.9. Pearson's Chi-squared test ($x^2$) is used to test the null hypothesis (no association). We conducted logistic regression under an additive genetic model to assess the association of all SNPs within the study with disease status of binary traits (0/1) for case and control subjects. Logistic regression association test is adjusted using Genomic Control (GC) to control population structure, the $p$-values are considered based on a GC inflation factor.

### 2.4 Logistic Regression

Logistic regression is defined as a statistical method for predicting binary outcomes [7]. Logistic regression modelling is a standard method commonly used for the analysis of genetic models. Let $Y \in \{0,1\}$ be a binary variable for disease status with 0 indicating a control and 1 indicating a case. Let $X \in \{0,1,2\}$ be a genotype at a particular SNP. Assuming that 0, 1, 2 represent homozygous major allele $AA$, heterozygous allele $Aa$ and homozygous minor allele $aa$ respectively. Logistic regression modelling is performed under an additive genetic model and it is given as [28]:

The conditional probability of $Y = 1$ is

$$\theta(X) = P(Y = 1|X) \quad (1)$$

The logit function which is the inverse of the sigmoidal logistic function is represented as

$$logit(X) = \ln \frac{\theta(X)}{1 - \theta(X)} \quad (2)$$

The logit is given as a linear predictor function as follows

$$logit(X) \sim \beta_0 + \beta_1 X \quad (3)$$

Logistic regression modelling is used to assess the association of all SNPs within the study with the phenotype. A p-value threshold of $10^{-2}$ is considered resulting in 6609 SNPs to help minimise the computational requirements need to process all SNPs and to remove less significant information. The final 6609 SNPs are used to train a DL stacked autoencoder and model the epistatic interactions between them. Information obtained from deep levels of the DL is used to initialize MLP models and fine-tuning for classification tasks. The trained classifier model is evaluated, and its predictive capacity assessed using cases and controls in T2D GWAS data.

### 2.5 Deep Learning

A multi-layer feedforward artificial neural network (MLP) is trained with batch gradient descent using back-propagation in this study for classification tasks, based on the theoretical definitions in [22], [24], [23]. The basic computational units of the network are neurons, for each neuron:

$$\begin{cases} x_1, x_2, \ldots x_n \text{ input for neuron} \\ a + 1 \text{ intercept term} \\ h_{W,b}(x) = f(W^T x) = f(\sum_{i=1}^{n} W_i x_i + b) \text{ output for neuron} \\ \text{where } f: \mathbb{R} \mapsto \mathbb{R} \text{ non linear activation function} \end{cases}$$

The non-linear activation function is a rectifier linear unit (ReLU) which computes a weighted sum of the inputs and is given according to:

$$f(x) = \max(0, x) \quad (4)$$

where $x$ denotes the input to the neuron.

The neural network is constructed using input, hidden, output layers containing a pre-defined a number of units (neuron) – depending on evaluation. The network representation model is given as:

$$\begin{cases} l \text{ is a layer} \\ n_l \text{ number of layers} \\ L_l \text{ a particular layer} \\ (W, b) = (W^{(1)}, b^{(1)}, W^{(n)}, b^{(n)}) \text{ parameters for} \\ \text{neural network} \\ W_{ij}^{(l)} \text{ weight of the connection between unit } j \text{ in layer } l, \\ \text{and unit } i \text{ in layer } l + 1 \\ b_i^{(l)} \text{ intercept node, bias of unit } i \text{ in layer } l + 1 \\ s_l \text{ number of units in layer } l \text{ (not including the bias unit)} \\ a_i^{(l)} \text{ is the activation of unit } i \text{ in layer } l \end{cases}$$

Given a fixed setting of parameters $W, b$ the neural network hypothesis is defined as $h_{W,b}(x)$ which gives the real number output.

The network is trained using a training subject $(x^{(i)}, y^{(i)})$ where $y^{(i)} \in \mathbb{R}^2$. The parameter $x$ is a vector of input features of a sample and $y$ is the outcome (in our case, a sample with T2D and sample without T2D). With a fixed training set $\{(x^{(1)}, y^{(1)}), \ldots, (x^{(m)}, y^{(m)})\}$ of $m$ training examples, the neural network can be trained using batch gradient descent and the following cost function [23]:

$$\begin{aligned} J(W, b) &= \left[\frac{1}{m} \sum_{i=1}^{m} J(W, b; x^{(i)}, y^{(i)})\right] \\ &+ \frac{\lambda}{2} \sum_{l=1}^{n_l-1} \sum_{i=1}^{s_l} \sum_{j=1}^{s_{l+1}} \left(W_{ji}^{(l)}\right)^2 \\ &= \left[\frac{1}{m} \sum_{i=1}^{m} \left(\frac{1}{2} \|h_{W,b}(x^{(i)}) - y^{(i)}\|^2\right)\right] \\ &+ \frac{\lambda}{2} \sum_{l=1}^{n_l-1} \sum_{i=1}^{s_l} \sum_{j=1}^{s_{l+1}} \left(W_{ji}^{(l)}\right)^2 \end{aligned} \quad (5)$$

where $J(W, b)$ is an average sum-of-squares error and the second term is a regularization term known as a weight decay term. The weight decay term helps to reduce the magnitude of the weights and prevent overfitting. Parameter $\lambda$ is the weight decay parameter and, it controls the relative importance of the first and second terms.

Before training the neural network model, the parameters $W_{ji}^{(l)}$ and each $b_i^{(l)}$ are random initialization to values close to zero. This step is essential to stop hidden layerunits learning the same function of the input. The gradient descent updates parameters $W, b$ as define below:

$$W_{ij}^{(l)} := W_{ij}^{(l)} - \alpha \frac{\partial}{\partial W_{ij}^{(l)}} J(W, b)$$

$$b_i^{(l)} := b_i^{(l)} - \alpha \frac{\partial}{\partial b_i^{(l)}} J(W, b)$$

(6)

where $\alpha$ represents the learning rate.

The partial derivatives of the cost function are computed using the backpropagation algorithm (see Algorithm 1).

The backpropagation algorithm first performs a feedforward pass to compute all the activations $a_i^{(l)}$ and the output value of $h_{W,b}(x)$ in the network. An error term $\delta_i^{(l)}$ is calculated for each node $i$ in layer $l$ to measure the contribution of this node to any errors in the output. For hidden nodes, the error term $\delta_i^{(l)}$ is computed using a weighted average $z_i^{(l)}$ of the error terms of the nodes that use $a_i^{(l)}$ as an input. For an output node, the error term $\delta_i^{(n_l)}$ (where $n_l$ is the output layer) signifies the difference between the network's activation and the true target value.

**Algorithm 1** Backpropagation Algorithm

1: Perform a feedforward pass and compute the activations for $L_2, L_3, \ldots, L_{n_l}$ ($n_l$ is the output layer)
2: **for** each output unit $i$ in layer $n_l$, **do**
3: $\delta_i^{(n_l)} = \frac{\partial}{\partial z_i^{(n_l)}} \frac{1}{2} \|y - h_{W,b}(x)\|^2$
$= -\left(y_i - a_i^{(n_l)}\right) \cdot f'(z_i^{(n_l)})$
4: **end for**
5: **for** $l = n_l - 1, \ldots, 2$, **do**
6:    **for** each node $i$ in layer $l$, **do**
7:      $\delta_i^{(l)} = \left(\sum_{j=1}^{s_{l+1}} W_{ji}^{(l)} \delta_j^{(l+1)}\right) f'(z_i^{(l)})$
8:    **end for**
9: **end for**
10: Compute partial derivatives:
11: $\frac{\partial}{\partial W_{ij}^{(l)}} J(W, b; x, y) = a_j^{(l)} \delta_i^{(l+1)}$
12: $\frac{\partial}{\partial b_i^{(l)}} J(W, b; x, y) = \delta_i^{(l+1)}$

Once the partial derivatives are computed, Equation (5), the derivative of the overall cost function $J(W, b)$ can be calculated as:

$$\frac{\partial}{\partial W_{ij}^{(l)}} J(W, b) = \left[\frac{1}{m} \sum_{i=1}^{m} \frac{\partial}{\partial W_{ij}^{(l)}} J(W, b; x^{(i)}, y^{(i)})\right] + \lambda W_{ij}^{(l)}$$

(7)

$$\frac{\partial}{\partial b_i^{(l)}} J(W, b) = \frac{1}{m} \sum_{i=1}^{m} \frac{\partial}{\partial b_i^{(l)}} J(W, b; x^{(i)}, y^{(i)})$$

Thereafter, batch gradient descent is used to train the neural network as described in Algorithm 2. $\Delta W^{(l)}$ is a matrix with dimensions similar to $W^{(l)}$, and $\Delta b^{(l)}$ is a vector with a similar dimension to $b^{(l)}$. Algorithm 2 describe the implementation of one iteration of batch gradient descent as follow:

**Algorithm 2** Batch Gradient Descent Algorithm

1: Set $\Delta W^{(l)} := 0, \Delta b^{(l)} := 0$ (matrix/vector of zeros) for all $l$.
2: **for** $i = 1, \ldots, m$, **do**
3: Use backpropagation to compute $\nabla_{W^{(l)}} J(W, b; x, y)$ and $\nabla_{b^{(l)}} J(W, b; x, y)$.
4: Set $\Delta W^{(l)} := \Delta W^{(l)} + \nabla_{W^{(l)}} J(W, b; x, y)$.
5: Set $\Delta b^{(l)} := \Delta b^{(l)} + \nabla_{b^{(l)}} J(W, b; x, y)$.
6: **end for**
7: Update the parameters:
8: $W^{(l)} := W^{(l)} - \alpha \left[\left(\frac{1}{m} \Delta W^{(l)}\right) + \lambda W^{(l)}\right]$
9: $b^{(l)} := b^{(l)} - \alpha \left[\frac{1}{m} \Delta b^{(l)}\right]$

The steps for batch gradient descent can be repeatedly implementing to minimize the overall cost function $J(W, b)$. Momentum training and learning rate annealing are advanced optimization tuning parameters that are used to modify back-propagation to allow previous iterations to influence the current version. The velocity vector is defined as follows:

$$v_{t+1} = \mu v_t - \alpha \nabla L(\theta_t)$$

$$\theta_{t+1} = \theta_t + v_{t+1}$$

(8)

where $\theta$ donates the parameters $W$ and $b$. The momentum coefficient is represented by $\mu$ and the learning rate is $\alpha$.

### 2.6 Stacked Autoencoders

Stacked autoencoders (SAE) are greedy layer wise unsupervised pretraining that apply autoencoders in each layer of a stacked network [29]. An autoencoder (AE) is an unsupervised learning method [30], [23] that can be used to train output values $\hat{x}$ to be similar to input values $x$ using backpropagation. The AE attempts to learn a function $h_{W,b}(x) \approx x$, which means it is trying to learn an approximation to the identity function and consequently output $\hat{x}$ such that is approximately equal to $x$. The aim is to discover interesting structure about the data specifically if there are correlations between the input features. This is achieved by limiting the number of hidden units in the network through the number of autoencoders used. In this case, the network will learn a compressed representation of the input, given the vector of hidden unit activations $a^{(2)} \in \mathbb{R}^n$, where $n$ is the number of hidden units, and then try to reconstruct the input $x$.

In autoencoders, most neurons are supposed to be in-active which means that the output values of these neurons are close to -1 when using a tanh activation function and close to 0 when a sigmoid activation function is considered. The activation of a hidden unit in the network is denoted as $a_j^{(2)}(x)$ for a given input $x$. Let

$$\hat{p}_j = \frac{1}{m}\sum_{i=1}^{m}\left[a_j^{(2)}(x^{(i)})\right] \quad (9)$$

signify the average activation of hidden unit $j$ and impose an approximation constraint $\hat{p}_j = p$, where $p$ is a sparsity parameter representing by a small value near to 0. To meet this constraint, the activation of the hidden unit must almost be close to 0. An extra penalty term that penalizes $\hat{p}_j$ is added, deviating significantly from $p$:

$$\sum_{j=1}^{s_2} p\, log\frac{p}{\hat{p}_j} + (1-p)log\frac{1-p}{1-\hat{p}_j} \quad (10)$$

where $s_2$ represents the number of units in the hidden layer, and $j$ is an index for summing the hidden units in the network. Kullback-Leibler (KL) divergence, which is a standard function used to measure how different two different distributions are, is used to impose the penalty term (10):

$$\sum_{j=1}^{s_2} KL(p||\hat{p}_j) \quad (11)$$

Where

$$KL(p||\hat{p}_j) = p\, log\frac{p}{\hat{p}_j} + (1-p)log\frac{1-p}{1-\hat{p}_j} \quad (12)$$

Equation (12) represents the Kullback-Leibler divergence between two Bernoulli random variables with mean $p$ and $\hat{p}_j$. And this can be equal to 0 if $\hat{p}_j = p$, alternatively it increases monotonically as $\hat{p}_j$ diverges from $p$. Hence, after adding a sparse penalty term, the overall cost function can now be defined as:

$$J_{sparse}(W,b) = J(W,b) + \beta \sum_{j=1}^{s_2} KL(p||\hat{p}_j) \quad (13)$$

where $J(W,b)$ defined previously in equation (5), and $\beta$ are used to control the sparsity penalty term's weight. Typically, the activations of hidden units are dependent on $W, b$ parameters, therefore the term $\hat{p}_j$ which is the average activation of hidden unit $j$ depends on $W, b$.

In order to integrate the KL-divergence term into the derivative calculation during the backpropagation algorithm (see Algorithm 1) the error term now is computed as:

$$\delta_i^{(l)} = \left(\left(\sum_{j=1}^{s_{l+1}} W_{ji}^{(l)} \delta_j^{(l+1)}\right) + \beta\left(-\frac{p}{\hat{p}_i} + \frac{1-p}{1-\hat{p}_i}\right)\right) f'(z_i^{(l)}) \quad (14)$$

This is a single layer autoencoder procedure. However, DL has better capability when several layers of autoencoders are stacked [29]. The concept of SAE is that the outputs of each layer are connected to the inputs of the subsequent layer, repeating this process for next layers, and for the classification task the last hidden layer is linked to a MLP classifier. After greedy layer wise unsupervised pretraining, the resulting deep features (epistatic interactions between SNPs) can be used as input to a deep supervised neural network. In this study, stacked autoencoder with three hidden layers of 2500, 1500, and 700 hidden neurons are used to pretrain the weights for the deep supervised neural network to classify individuals with and without T2D.

### 2.7 Performance Measures

The performance of the proposed DL stacked autoencoder classifier is measured using the Area Under the Curve (AUC), Sensitivity, Specificity, Gini, Logarithmic Loss, and MSE values. The dataset is split randomly into training (80%), validation (10%), and testing (10%).

Sensitivity and specificity are used to measure the positive and negative predictive capabilities of classifiers in binary classification. Sensitivity refers to the true positive rate and describes the ability of the test to correctly classify people with T2D. While Specificity describes the true negative rate which is the ability of the test to correctly classify people without T2D [31].

The area under the curve (AUC) and the receiver operating characteristic curve (ROC curve) are both used to assess and compare the performance of different classifier models and are commonly used in binary classification studies [32]. AUC represents the probability of a correct classification for positive and negative instances; the positive class will be ranked higher thus a higher AUC means a better classification [32]. While ROC curve is a graphical plot to display the performance of a binary classification model. It is created by plotting the true positive rate (also known as sensitivity) against the false positive rate which can be represented as (1-specificity) [32].

The Gini coefficient value can be derived from the AUC of the ROC curve where $Gini = 2 * AUC - 1$. It represents the area between the ROC curve and the diagonal. The Gini coefficient is usually used in binary classification problems. A Gini value above 60% is considered a good model.

Logarithmic Loss (Logloss) is a classification loss function often used to measure the performance of a classification model where the prediction input is a probability value between 0 and 1. Logloss increases as predicted probability (accuracy) decreases. A Logloss value of 0 is an indication of a perfect model where the model correctly classifies all class instances.

The Mean Squared Error (MSE) performance metric is used to measure the average of the squares of the errors which is the difference between actual values and the predicted values. MSE values close to 0 mean that the model correctly classifies all class instances.

DL stacked autoencoder (unsupervised learning) is used to find the epistatic interactions between SNPs before the final hidden layer is used to initialize and fine-tune an MLP for binary classification of T2D and is benchmarked with the supervised DL and RF classifier models. In the case of the RF, the number of trees is set to 400 with a maximum tree depth of 40 for training. For the DL a RectifierWithDropout activation function is used with 100 epochs and four hidden layers with 10 neurons in each layer. Input dropout ratio is set to 0.1 and hidden dropout ratios for each layer set to 0.5. Early stopping is adopted using a stopping metric set to logloss, with stopping tolerance and stopping rounds set to $1 \times 10^{-2}$ and 5 respectively. The learning rate configured to 0.005 with rate annealing, and rate decay set to $1 \times 10^{-6}$ and 1 respectively. Momentum start is set to 0.5 with momentum stable to 0 and momentum ramp to $1 \times 10^6$. The max w2 is set to 10.

For the first stacked autoencoder models (2500 hidden units), a RectifierWithDropout activation function is used, and the number of epochs set to 10 iterations with two hidden layers containing 10 neurons each. For the second (2500, 1500 hidden neurons) and third (2500,1500,700 hidden neurons) models, a RectifierWithDropout activation function is used, with 10 epochs and two hidden layers with 20 neurons each. For all stacked autoencoder classifiers an adaptive learning rate is adopted with parameters *rho* and *epsilon* set to 0.99 and $1 \times 10^{-8}$ respectively. Table 1 provides the description of tuning parameters used with the DL classifier model.

## 3 RESULTS

This section presents the classification results for T2D obtained using the proposed SAE approach. In comparison, the DL and RF classifiers are used to benchmark the performance of the SAE. This evaluation considers SNPs generated with a p-value threshold of $10^{-2}$ resulting in 6609 SNPs. The stacked autoencoder uses these SNPs to extract the non-linear epistatic interactions between SNPs. The first SAE consists of 2500 hidden neurons while the second and third SAE use (2500, 1500), and (2500, 1500, 700) hidden neurons respectively.

Table 2 illustrates the performance metrics for the SAE, DL, and RF using the validation set. Metric values for the first SAE (2500 hidden units), second SAE (2500,1500), and third SAE (2500,1500,700) were obtained using optimized F1 threshold with values 0.3412, 0.3192, and 0.3474 respectively. The performance metrics for DL and RF were acquired using optimized F1 threshold with values 0.37, and 0.5 respectively.

TABLE 1
DEFINITION OF TUNING PARAMETERS USED WITH DL

| Tuning Parameter | Description of Tuning Parameter |
|---|---|
| Input dropout ratio | A fraction of the features for each training row to be removed from training. This can improve generalization. |
| Hidden dropout ratios | A fraction of the inputs for each hidden layer to be removed from training. This can improve generalization. |
| Stopping metric | Is used to determine the metric to use for early stopping |
| Stopping tolerance | Is used to set the relative tolerance for the metric-based stopping to stop training in case the improvement is less than tolerance value. |
| stopping rounds | Is used to stop training if the option selected for stopping metric doesn't improve for the specified value of training rounds. |
| Learning rate | Is a function of the difference between the predicted value and the target value. Back propagation is used to correct the output at each hidden layer. |
| Rate annealing | Is used to reduce the learning rate to freeze into local minima. |
| Rate decay | Is used to control the change of learning rate throughout layers. |
| Momentum start | Is used to control the amount of momentum at the beginning of training. |
| Momentum stable | Is used to control the amount of learning for which momentum increases |
| Momentum ramp | Is used to control the final momentum value reached after momentum ramp training samples |
| Max w2 | Is a maximum on the sum of the squared incoming weights into any one neuron. It is useful when activation function is set to Rectifier. This help stability for Rectifier. |

TABLE 2
PERFORMANCE METRICS FOR SAE, DL, RF FOR VALIDATION SET

| Models | AUC | Sens | Spec | Logloss | Gini | MSE |
|---|---|---|---|---|---|---|
| SAE1 | 0.9546 | 0.9309 | 0.8846 | 0.3241 | 0.9093 | 0.0778 |
| SAE2 | 0.9206 | 0.9090 | 0.7863 | 0.3932 | 0.8413 | 0.1129 |
| SAE3 | 0.8689 | 0.8618 | 0.7264 | 0.4710 | 0.7379 | 0.1501 |
| DL | 0.9781 | 0.9709 | 0.8888 | 0.2878 | 0.9562 | 0.0812 |
| RF | 0.7436 | 0.9163 | 0.2863 | 0.6519 | 0.4872 | 0.2297 |

*SAE1=2500 hidden units.*
*SAE2=2500,1500 hidden units.*
*SAE3=2500,1500,700 hidden units.*

Table 3 presents the performance metrics obtained using the test set for the SAE, DL, and RF models. Metric values for the first SAE (2500 hidden units), second SAE (2500,1500), and third SAE (2500,1500,700) were obtained using optimized F1 thresholds with values 0.2770, 0.4340, and 0.3424 respectively. The performance metrics for the DL and RF were gained using optimized F1 thresholds with values 0.3683, and 0.515 respectively. The results are lower than those produced using the validation set except for the third SAE (2500, 1500, 700) which showed a 2.49% improvement.

TABLE 3
PERFORMANCE METRICS FOR SAE, DL, RF FOR THE TEST SET

| Models | AUC | Sens | Spec | Logloss | Gini | MSE |
|---|---|---|---|---|---|---|
| SAE1 | 0.9289 | 0.9087 | 0.8053 | 0.4582 | 0.8578 | 0.1162 |
| SAE2 | 0.8950 | 0.8547 | 0.7977 | 0.4761 | 0.7901 | 0.1391 |
| SAE3 | 0.8938 | 0.9020 | 0.6984 | 0.4155 | 0.7877 | 0.1334 |
| DL | 0.9674 | 0.9628 | 0.8435 | 0.3117 | 0.9349 | 0.0893 |
| RF | 0.7324 | 0.8310 | 0.4656 | 0.6553 | 0.4649 | 0.2313 |

The classification accuracy of SAE shows a progressive deterioration as the input features are steadily compressed down to 700 hidden neurons using validation and test sets. DL classifier achieved comparable results to those produced using SAE (2500 compressed units) in the validation set. The results evidently show that SAE outperforms the RF classifier.

Fig.2 presents the ROC curves for the SAE, DL and RF classifiers. Despite the gradual deterioration in the performance of SAE, satisfactory results are achieved with 700 hidden units.

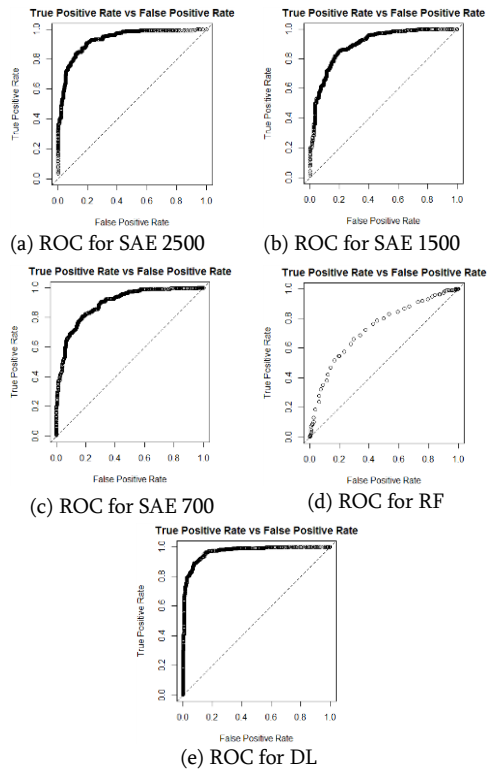

(a) ROC for SAE 2500  (b) ROC for SAE 1500
(c) ROC for SAE 700   (d) ROC for RF
(e) ROC for DL

Fig. 1. Performance ROC curves for the test set. (a) to (c) for SAE. (d) for RF. (e) for DL.

## 4 DISCUSSION

Genetic association studies have significantly expanded our understanding of the genetic variants that predispose us to human diseases. However, in complex disease, such as T2D, using a standard statistical test for single-SNP analysis has proven ineffective as single genetic loci (SNP) do not act independently to increase disease risk. GWAS fail to find the non-linear relationships that exist between SNPs and genotype-phenotype interactions. Standard multi-variable statistical approaches, such as logistic regression, are more suited to capturing linear interactions not epistatic interactions.

Using SAE, the results in this paper the results show a gradual deterioration in performance as the number of features is compressed down to 700 hidden units. However, the classification accuracy value of the 700 compressed neurons remains reasonable with 89.38% in the test set. The best result obtained using 2500 compressed units (AUC=92.89%, Sens=90.87%, Spec=80.53%, Logloss= 45.82%, Gini=85.78%, MSE=11.62%).

RF and DL algorithms were used to benchmark the SAE classification performance. RF is a prevalent method used in genetic studies [13], [14], [33], [34]. In this analysis, the result shows that using 6609 SNPs it was possible to achieve 73.24% classification accuracy. Sensitivity and specificity are instable indicating that RF classifier has the low discriminatory capacity for this given dataset to separate cases and controls phenotypes. In comparison, even though SAE showing trivial deterioration ranging from AUC=92.89% to AUC=89.38% the results remain significantly higher than those achieved by the RF model. This is because, in stacked autoencoders, the multiple hidden layers compress the input features into abstract representations that capture and model the complexity of non-linearity found within genotype-phenotype interactions observed in genetic data. This unsupervised learning algorithm outperforms the traditional supervised classification models and offers a powerful way to enhance GWAS data analysis.

Using 6609 SNPs to train the DL model it was possible to obtain (AUC=96.74%, Sens=96.28%, Spec=84.35%, Logloss= 31.17%, Gini=93.49%, MSE=8.93%). Although SAE with 2500 compressed units (initially 6609 SNPs) achieved less predictive accuracy than the DL model using (6609 SNPs) the results are comparable and significant for both models with AUC=92.89% for SAE and AUC=96.74% for DL. The fact that we still obtain high results even though the original data is compressed from 6609 SNPs to 700 SNPs with a lower predictive accuracy of 89.38% is encouraging and demonstrates the potential of applying DL pre-initialized classification models with stacked autoencoder for the classification of type 2 diabetes GWAS data.

This paper presents a novel framework for the classification of T2D case-control GWAS data. The combination of an unsupervised learning approach using SAE and subsequent fine-tuning for a supervised DL model to extract latent representation from large scale biological data structure shows potential. This kind of framework for the classification of GWAS data particularly T2D genomic data could provide a starting point for researchers and professionals investigating the aetiology of T2D and could answer the question of the remaining missing heritability that traditional statistical approach fails to explain.

## 5 CONCLUSION

In this paper, a novel framework for the classification of T2D genetic data is proposed. We investigated the potential of using a deep learning stacked autoencoder for de-tecting epistatic interactions and performing classification tasks using high-dimensional GWAS data in T2D. This study used existing datasets provided by the Genotypes and Phenotypes (dbGap) database. Various stringent quality control assessment steps followed by logistic regression association analysis adjusted GC are performed for single-SNP analysis. Using 5393 T2D case-control samples, we achieved (AUC=89.38%, Sens=90.20%, Spec=69.84%, Logloss= 41.55%, Gini=78.77%, MSE=13.34%) using 700 compressed units.

Despite the encouraging results produced in this paper, more in-depth investigations are still required. There are many possible areas for improvement in our approach; an important one includes parameter tuning and optimization for DL algorithms in an attempt to further enhance classification results.

The fact that we use a p-value threshold to extract a subset of SNPs potentially excludes variables with small effects individually but may contribute much larger within clusters on SNPs.

Overall, the proposed methodology is robust and contributes to the bioinformatics and computational biology fields and provides new insights into the potential use of unsupervised learning algorithms over supervised traditional approaches when analyzing high-dimensional GWAS data that we believe warrants further investigation.

ACKNOWLEDGMENT

The dataset(s) used for the analyses described in this manuscript were obtained from the database of Genotype and Phenotype (dbGaP) found at http://www.ncbi.nlm.nih.gov/gap through dbGaP accession number phs000091.v2.p1. The Nurses' Health Study (NHS) and Health Professionals' Follow-up Study (HPFS) is part of the Gene Environment Association Studies initiative (GENEVA, http://www.genevastudy.org) funded by the trans-NIH Genes, Environment, and Health Initiative (GEI).